\documentclass[preprint,12pt,authoryear]{elsarticle}

\usepackage{amssymb}
\usepackage{booktabs} 
\usepackage[table]{xcolor} 
\usepackage{amsmath}  
\usepackage{stfloats}
\usepackage{subfig}
\usepackage{graphicx}
\usepackage{hyperref}

\begin{document}

\begin{frontmatter}



\title{CNN-Based Classification of Persian Miniature Paintings from Five Renowned Schools}


\author[QUT]{Mojtaba Shahi}
\author[UniGe]{Roozbeh Rajabi}
\author[AUI]{Farnaz Masoumzadeh}

\affiliation[QUT]{organization={Qom University of Technology},
			city={Qom},
			country={Iran}}

\affiliation[UniGe]{organization={University of Genoa},
            city={Genoa},
            country={Italy}}
            
\affiliation[AUI]{organization={Art University Of Isfahan},
           	city={Isfahan},
           	country={Iran}}

\begin{abstract}
This article addresses the gap in computational painting analysis focused on Persian miniature painting, a rich cultural and artistic heritage. It introduces a novel approach using Convolutional Neural Networks (CNN) to classify Persian miniatures from five schools: Herat, Tabriz-e Avval, Shiraz-e Avval, Tabriz-e Dovvom, and Qajar. The method achieves an average accuracy of over 91\%. A meticulously curated dataset captures the distinct features of each school, with a patch-based CNN approach classifying image segments independently before merging results for enhanced accuracy. This research contributes significantly to digital art analysis, providing detailed insights into the dataset, CNN architecture, training, and validation processes. It highlights the potential for future advancements in automated art analysis, bridging machine learning, art history, and digital humanities, thereby aiding the preservation and understanding of Persian cultural heritage.
\end{abstract}




\end{frontmatter}


\section{Introduction}
\begin{figure}[t]
	\centering
	\includegraphics[width=0.5\textwidth]{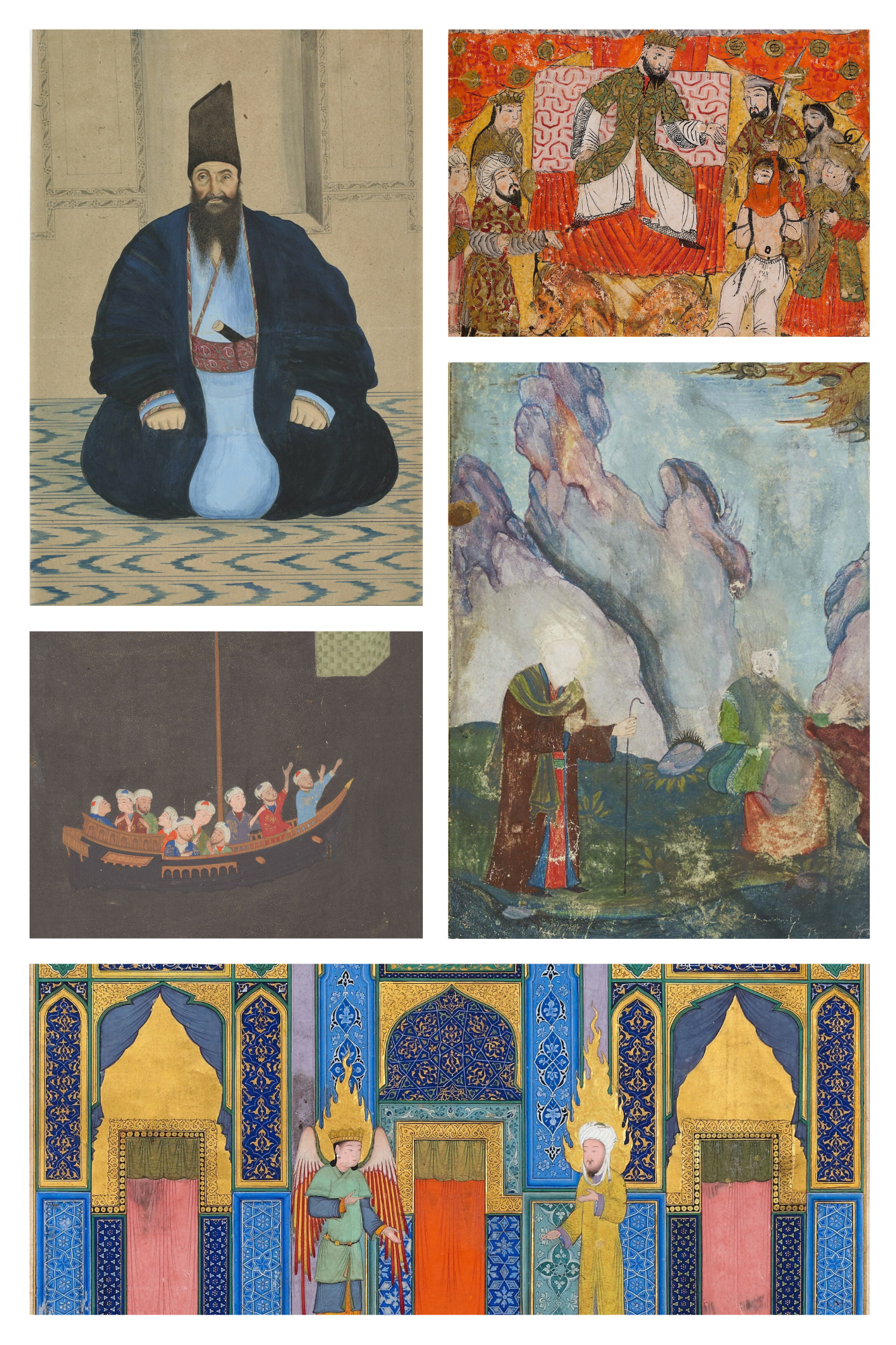}
	\caption{An exquisite collage of Persian miniatures, showcasing a rich tapestry of styles from various schools of art, each piece a testament to the diverse cultural narratives and artistic expressions of the Persian heritage.}
	\label{fig:collage-mix}
\end{figure}

The fusion of art and technology has opened new avenues in the study and appreciation of visual arts, leading to a renaissance in computational painting analysis. This interdisciplinary approach has significantly enhanced our understanding of various art forms, yet certain areas remain less explored. Among these is the rich tradition of Persian miniature painting, a genre that holds a unique position in the annals of art history due to its distinctive styles and cultural significance. Persian miniatures, characterized by their intricate designs and vivid storytelling, are more than mere artistic expressions; they are a window into the historical, cultural, and social fabric of their times. These paintings are distinguished not only by their aesthetic appeal but also by their deep symbolic meanings and the portrayal of contemporary life, making them invaluable for cultural studies. As illustrated in Figure \ref{fig:collage-mix}, the diversity and richness of Persian artistic heritage are vividly captured through a carefully curated collage of Persian miniatures, highlighting the varied schools of art and cultural narratives.

The schools of Persian miniatures, each a distinct universe of style and expression, are closely tied to the dynasties under which they flourished. For instance, the Tabriz school, prospering during the Safavid era, is renowned for its vivid colors and dynamic scenes, reflecting the opulence and grandeur of Safavid rule. In contrast, the Shiraz school, with its romantic and emotional style, often depicted in soft hues, illustrates the cultural milieu of the region, marked by poetic landscapes and love stories \citep{loukonine2010persian}.

The geographical diversity has also played a crucial role in shaping these schools. The Isfahan school, emerging as the artistic capital during the Safavid period, is distinguished by its balanced and harmonious compositions, blending miniature painting with calligraphy, a reflection of Isfahan's status as a cultural hub. Similarly, the Qazvin school, known for its bold color schemes and aesthetic emphasis, represents the artistic evolution under the Safavid dynasty's patronage. Moreover, Persian miniatures are a canvas where eastern and western influences converge. The impact of Chinese art is evident in the Herat school, particularly during the Timurid era, where elements like dragons and phoenixes, along with specific color palettes, are prominent. The Qajar school, on the other hand, reveals the influence of Western art, evident in its realistic portraiture and adoption of European painting techniques, highlighting the cultural exchanges during the Qajar dynasty \citep{o2021studies}.

\begin{figure}[t]
	\centering
	\includegraphics[width=0.5\textwidth]{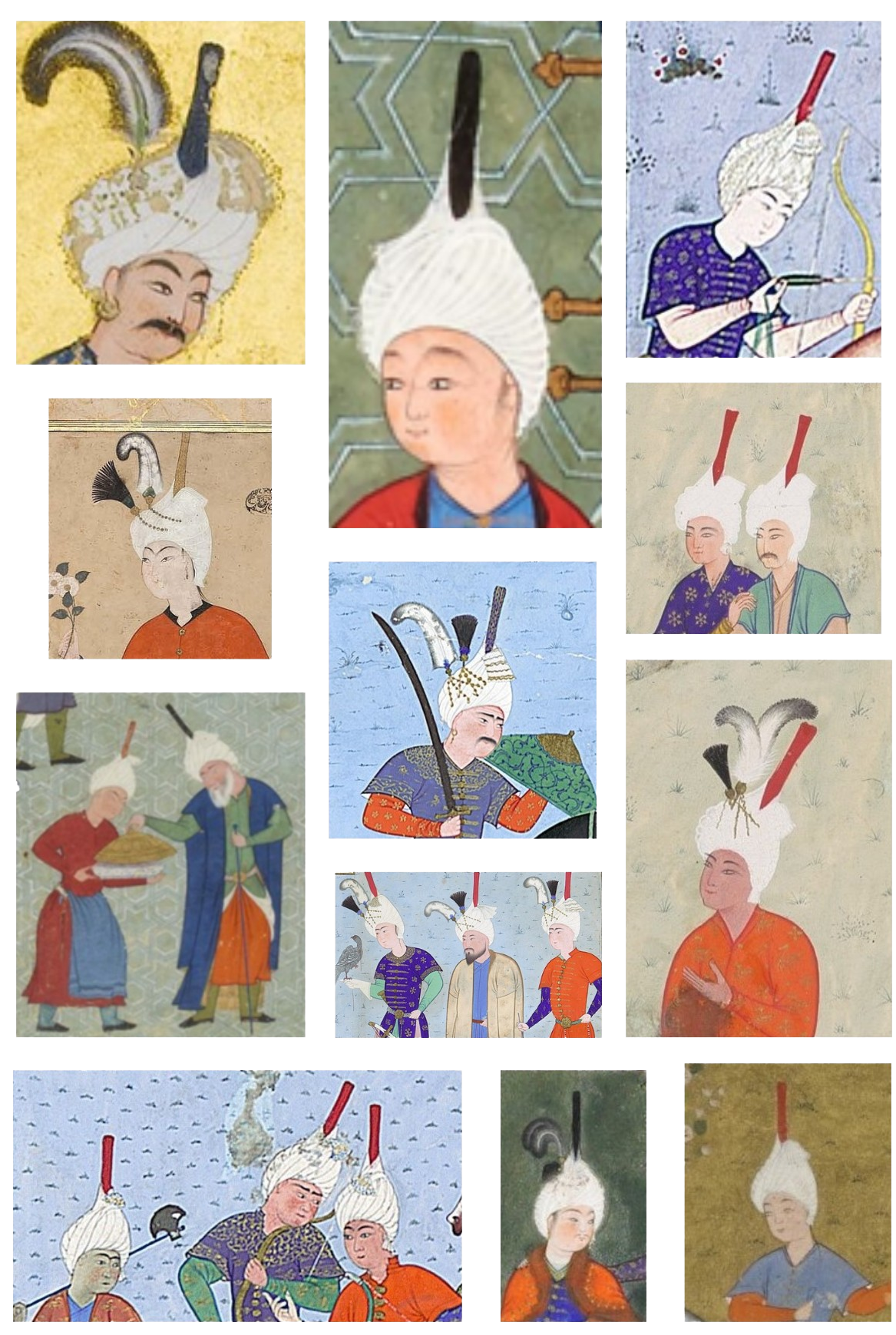}
	\caption{An illustration of the Qizilbash hat within the Persian miniatures of the Tabriz School.}
	\label{fig:Qizilbash}
\end{figure}

In exploring the distinctive features of Persian miniatures that allow for the differentiation of various schools within this wealthy artistic tradition, it is imperative to acknowledge the intricate symbols and motifs that carry profound cultural and historical significance. A notable example is the emblematic Qizilbash hat, which serves as a unique identifier of the Tabriz School during the Safavid dynasty. This highlights how elements such as the Qizilbash hat not only denote religious and political affiliations but also subtly convey narratives of power, loyalty, and societal order within the artwork. Figure \ref{fig:Qizilbash} demonstrates the incorporation of the Qizilbash hat in the Tabriz school's Persian miniatures, reflecting how art intricately reflects the rich tapestry of cultural, spiritual, and political influences of its era\citep{qizilbash_hat}.

In the approach of this paper, we leverage the capabilities of pre-trained Convolutional Neural Networks (CNNs) for feature extraction from the images of Persian miniatures. Recognizing the unique and intricate details inherent in these artworks, CNNs offer a robust framework for capturing the nuanced visual elements that define each school of art. Upon extracting these features, we introduce shallow, fully connected layers tailored to the specific task of classifying these artworks into five distinct classes corresponding to the major schools of Persian miniature painting. Our research contributes significantly to the field of digital art analysis by presenting a meticulously curated dataset that encapsulates the distinct features of each school. This dataset not only serves as a foundation for our CNN-based methodology but also as a valuable resource for further studies in art history and digital humanities. Through this work, we aim to enhance the understanding and preservation of Persian miniatures, offering new perspectives and tools for art historians, curators, and enthusiasts alike. 

The remainder of this paper is structured as follows: Section \ref{sec:relatedworks}, provides a comprehensive review of previous research in the field of art style classification, highlighting the advances and identifying the gaps that our study aims to fill. In Section \ref{sec:methodology}, we delve into the core of our research approach, beginning with patch extraction, where we describe the process of segmenting Persian miniature images into patches, and transfer learning, outlining how we adapt a pre-trained neural network for the classification of these patches. Section \ref{sec:experimentalsetup}, details the practical aspects of our research. This includes dataset, where we describe the unique collection of Persian miniatures used in our study, classification models, which explains the architecture and functioning of the neural network model we employ, and evaluation metrics, where we discuss the criteria used to assess the performance of our model. In Section \ref{sec:results}, we present the findings of our experiments, analyzing and interpreting the results in context of Persian miniature classification. Finally Section \ref{sec:conclusion}, summarizes our study, reflecting on its contributions, limitations, and potential areas for future research.

\section{Related Works}
\label{sec:relatedworks}

One of the early approaches in this area was presented in a study in 2009. The authors introduced an automatic method for classifying digital images of paintings into five artistic styles or genres, a task traditionally reserved for human experts. They utilized both color and edge features of the images and employed a decision tree algorithm. Their evaluation was based on a diverse dataset of 353 digital images gathered from various Internet sources, covering genres such as Abstract Expressionism, Cubism, Impressionism, Pop Art, and Realism. This diversity aimed to reflect the quality variation typically found in consumer-grade digital captures of paintings. Despite the challenges posed by the variable resolution and quality of the dataset, they achieved a classification accuracy of 69.1\%. This research marked a significant step towards the automation of art classification, proving the efficacy of machine learning in tasks traditionally dominated by subjective human interpretation \citep{zujovic2009classifying}.

A subsequent study in 2010 focused on classifying artworks of seven different artists using a multi-class Support Vector Machine (SVM) with advanced features. The study specifically targeted paintings by Cezanne, Dali, Durer, Monet, Picasso, Rembrandt, and Van Gogh, obtaining around 200 images per artist for a total of over 1400 training examples through a script that downloaded images from Google Images. The initial feature selection included pixels of images and color histograms, with each image normalized to a 100x100 pixel dimension and the color space reduced to 18-bit depth. However, the accuracy of these basic features was not sufficient, prompting the exploration of more advanced features inspired by previous research \citep{xiao2010sun}. These included GIST \citep{oliva2001modeling}, Histogram of Oriented Gradients (HOG) \citep{dalal2005histograms}, Dense SIFT \citep{lazebnik2006beyond}, Local Binary Patterns (LBP) \citep{ahonen2009rotation}, Sparse SIFT histograms \citep{matas2004robust} - \citep{sivic2004video}, SSIM \citep{shechtman2007matching}, thumbnail images \citep{torralba200880}, line features \citep{kovsecka2002video}, textural histograms \citep{martin2001database}, color histograms, geometric probability maps \citep{hoiem2007recovering}, and geometric histograms \citep{lalonde2007photo}. Multiple kernels for each feature set, including the $\chi^2$ kernel, linear kernel, histogram intersection kernel, and radial basis function, were evaluated. By weighting the performance of different feature sets, a combination of these features resulted in an improved accuracy of 85.3\% for the multi-class problem, where the goal was to correctly identify the artist from one of the seven different classes. This study underscored the potential of machine learning in art classification and highlighted the importance of feature selection and combination in achieving high accuracy, as well as the significance of having a diverse and representative dataset \citep{blessing2010using}.

In another notable work in 2014, researchers employed Restricted Boltzmann Machines and image processing techniques for feature extraction and used a genetic algorithm to optimize the weights for a k-nearest neighbors algorithm in classifying paintings by three different artists, achieving an accuracy of 90.44\%. This analysis was conducted on a dataset consisting of 120 digital reproductions of paintings by Rembrandt, Renoir, and Van Gogh, downloaded from the Webmuseum. The digital reproductions ensured uniform acquisition across the images, allowing classification to be based on artist characteristics rather than digital artifacts, thereby demonstrating the effectiveness of combining traditional machine learning algorithms with evolutionary optimization techniques for art classification \citep{levy2014genetic}.

A significant leap was made in 2016, where a study utilized an autoencoder for feature extraction and a Convolutional Neural Network (CNN) for model training, achieving an impressive accuracy of 96.52\%. This research highlighted the power of deep learning techniques in extracting intricate patterns and features from artworks. The study was conducted using a dataset of 5,000 paintings from the Webmuseum, with images having 24-bit color depth and resolutions averaged approximately at 1000 × 1000 pixels, normalized to 256×256 pixels for the training. The supervised classification benchmark utilized in their experiments consisted of digital reproductions of 120 paintings by Rembrandt, Renoir, and Van Gogh, demonstrating the effectiveness of convolutional autoencoders in painter classification on a diverse and challenging dataset \citep{david2016deeppainter}.

In a more recent study in 2019, researchers proposed a two-stage method for classifying the style of art paintings. Initially, images were divided into five parts, and each part was classified separately using a deep convolutional neural network. In the next step, the decision was made using a probability vector derived from the classified parts as input to another neural network. The best result for this scenario was obtained when an additional sixth part, the entire but normalized image, was considered. This approach was applied to three datasets, including images of fine art paintings from public and private collections, with 6, 29, and 19 classes, respectively, achieving accuracies of 67.16\%, 66.71\%, and 77.53\%. Specifically, Dataset 1 included 30,870 images representing six artistic styles, Dataset 2 comprised 26,400 images across 22 styles, and Dataset 3 contained 19,320 images classified into 19 styles. The experiments indicated significant improvements over existing baseline techniques, utilizing the pre-trained InceptionV3 deep CNN as the classifier for image parts in the first stage \citep{sandoval2019two}.

\section{Methodology}
\label{sec:methodology}

\subsection{Patch Extraction}

\begin{figure}[t]
	\centering
	\includegraphics[width=0.75\textwidth]{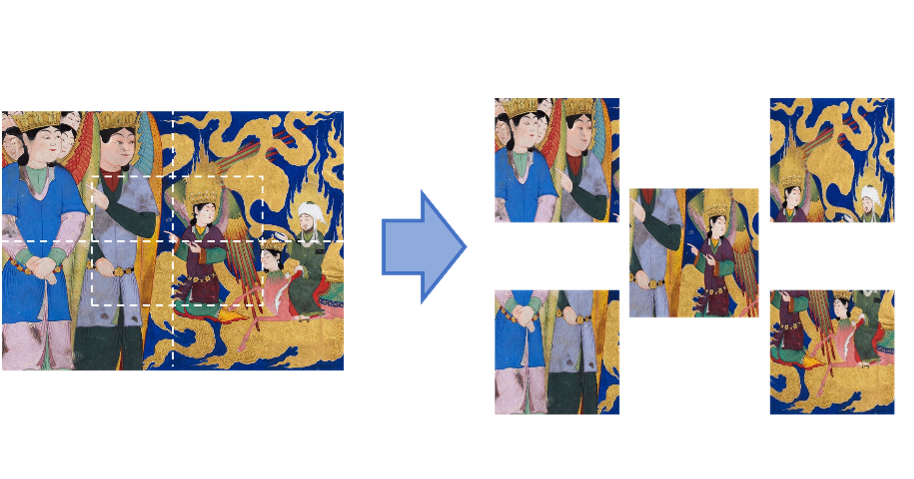}
	\caption{An illustration of artwork segmented into five normalized patches, showing four non-overlapping corners and a central patch overlapping each by a quarter.}
	
	\label{fig:decompose}
\end{figure}

Inspired by the approaches utilized in previous studies \citep{sandoval2019two} - \citep{folego2016impressionism}, our research adopts a specific technique for generating input data. This involves dividing each image into five distinct patches. This process aims for more detailed analysis and extraction of extensive features from every segment of the images. Each of these five patches includes four pieces from the image's corners and one central piece, ensuring that all critical aspects of the image are considered. After dividing the images, each patch is normalized separately. The normalization process involves resizing all patches to 256x256 pixels and adjusting their pixel values based on a standard scale. Figure \ref{fig:decompose} presents an example from our dataset. This step ensures the consistency and standardization of the input data. Moreover, each patch is labeled using one-hot encoding to inject the input data more accurately into the machine learning models. This approach not only aids in enhancing the quality and precision of the analysis but also leads to better identification and classification of artistic schools in Persian miniature works.

\subsection{Transfer Learning}
In our research, we have embraced the Transfer Learning methodology for feature extraction, with a focus on employing pre-trained models. This approach is augmented by a novel application of the random dropout technique, enhancing the model's ability to generalize and prevent overfitting. The pre-trained models are configured with their weights frozen to maintain the integrity of their learned features \citep{yosinski2014transferable}. To customize these models for our specific task, new layers are added. These include layers designed for pooling and dense connections, along with dropout layers. The dropout layers randomly deactivate a subset of neurons during training, which helps in reducing over-reliance on certain paths within the network, thereby encouraging a more distributed and robust learning \citep{srivastava2014dropout}. The architecture is completed with a final classification layer, tailored to the specific requirements of our dataset. This model setup, especially with the integration of dropout, strategically navigates the challenges of training complexity and model overfitting, aiming to achieve a more reliable and nuanced understanding of our data.

\begin{figure*}[t]
	\centering
	
	\subfloat[]{
		\includegraphics[width=0.25\textwidth]{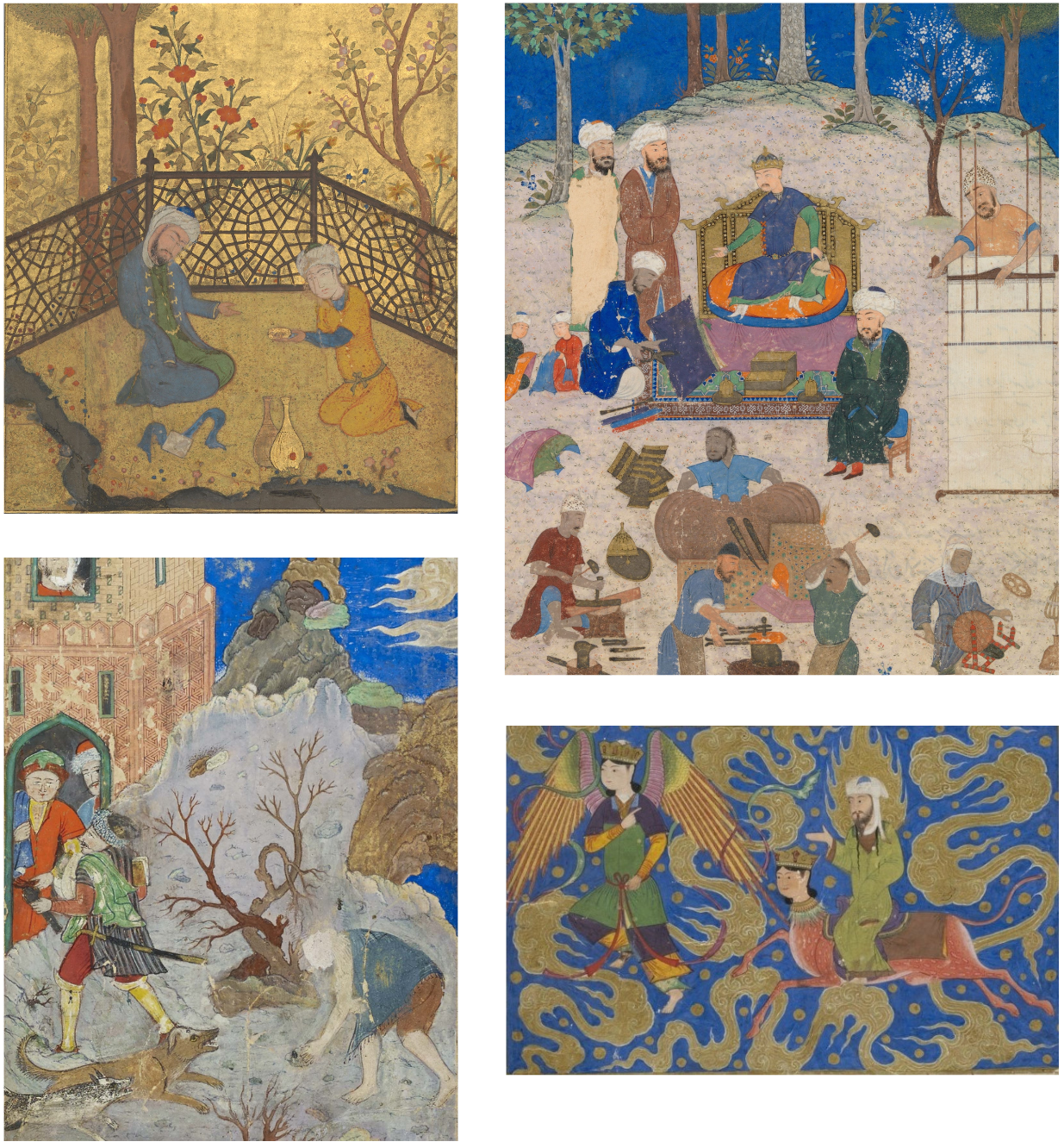}
		\label{fig:herat}
	}
	\hfill
	\subfloat[]{
		\includegraphics[width=0.3\textwidth]{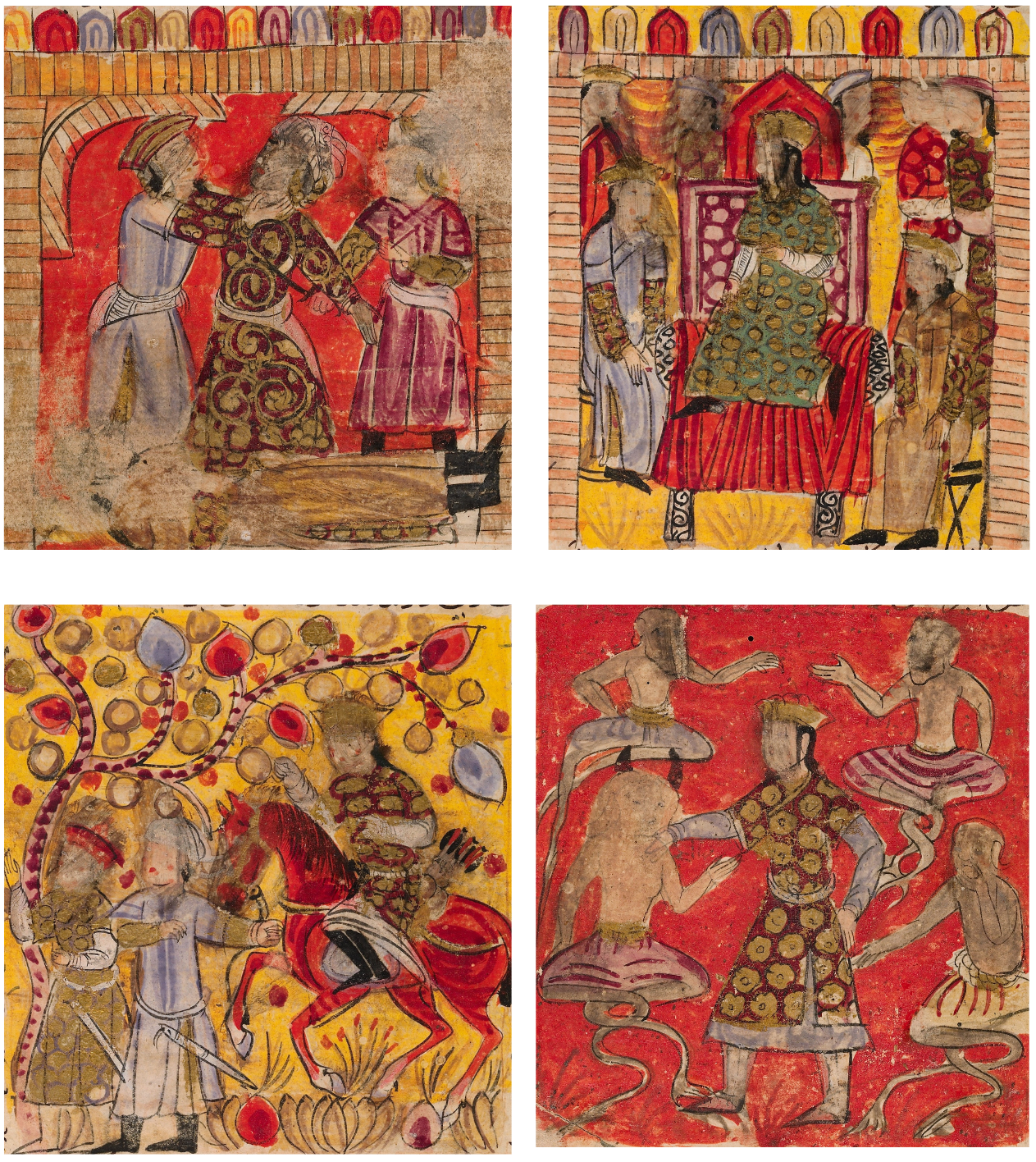}
		\label{fig:sh_1}
	}
	\hfill
	\subfloat[]{
		\includegraphics[width=0.3\textwidth]{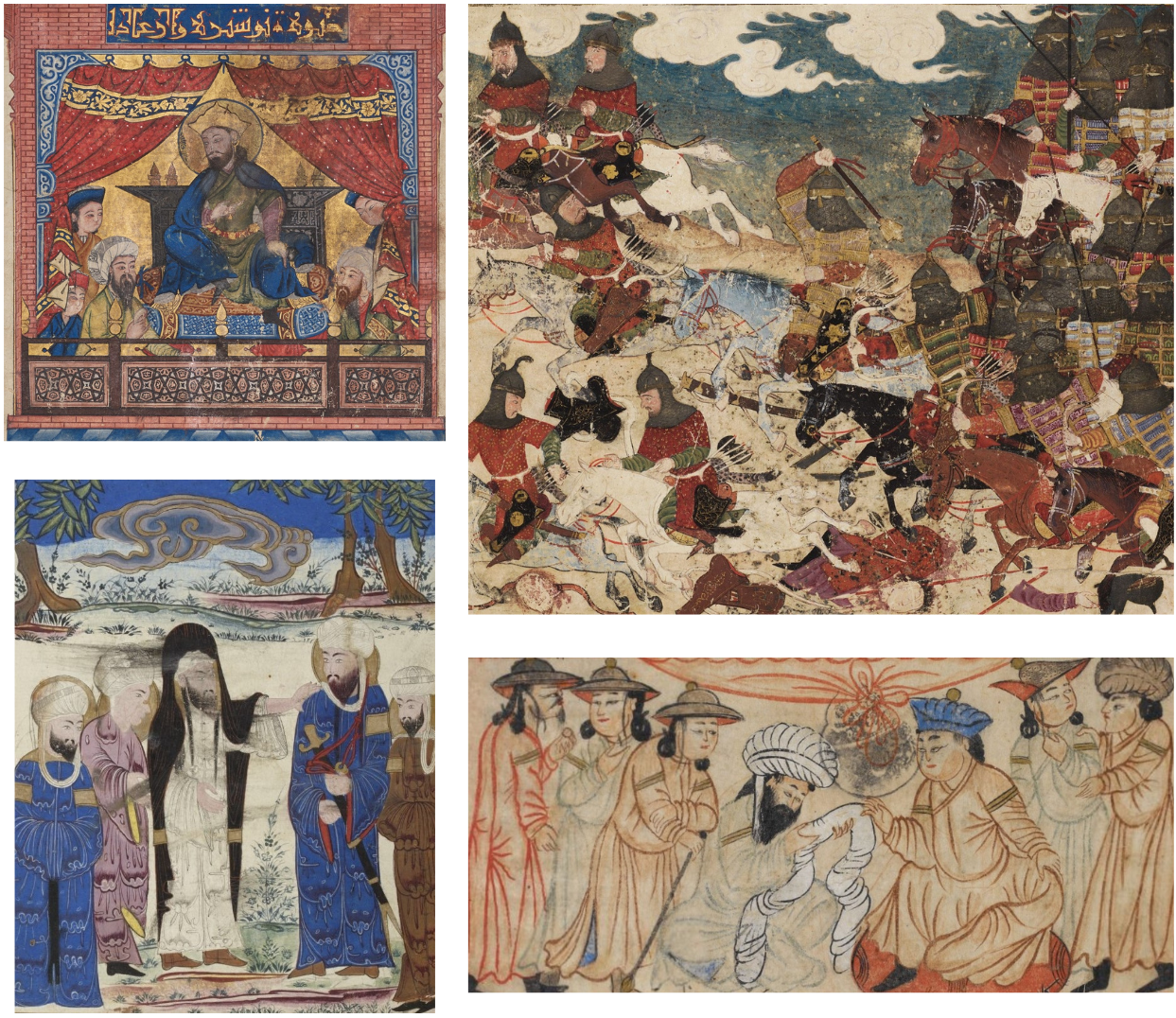}
		\label{fig:tabriz_1}
	}
	\hfill
	\subfloat[]{
		\includegraphics[width=0.5\textwidth]{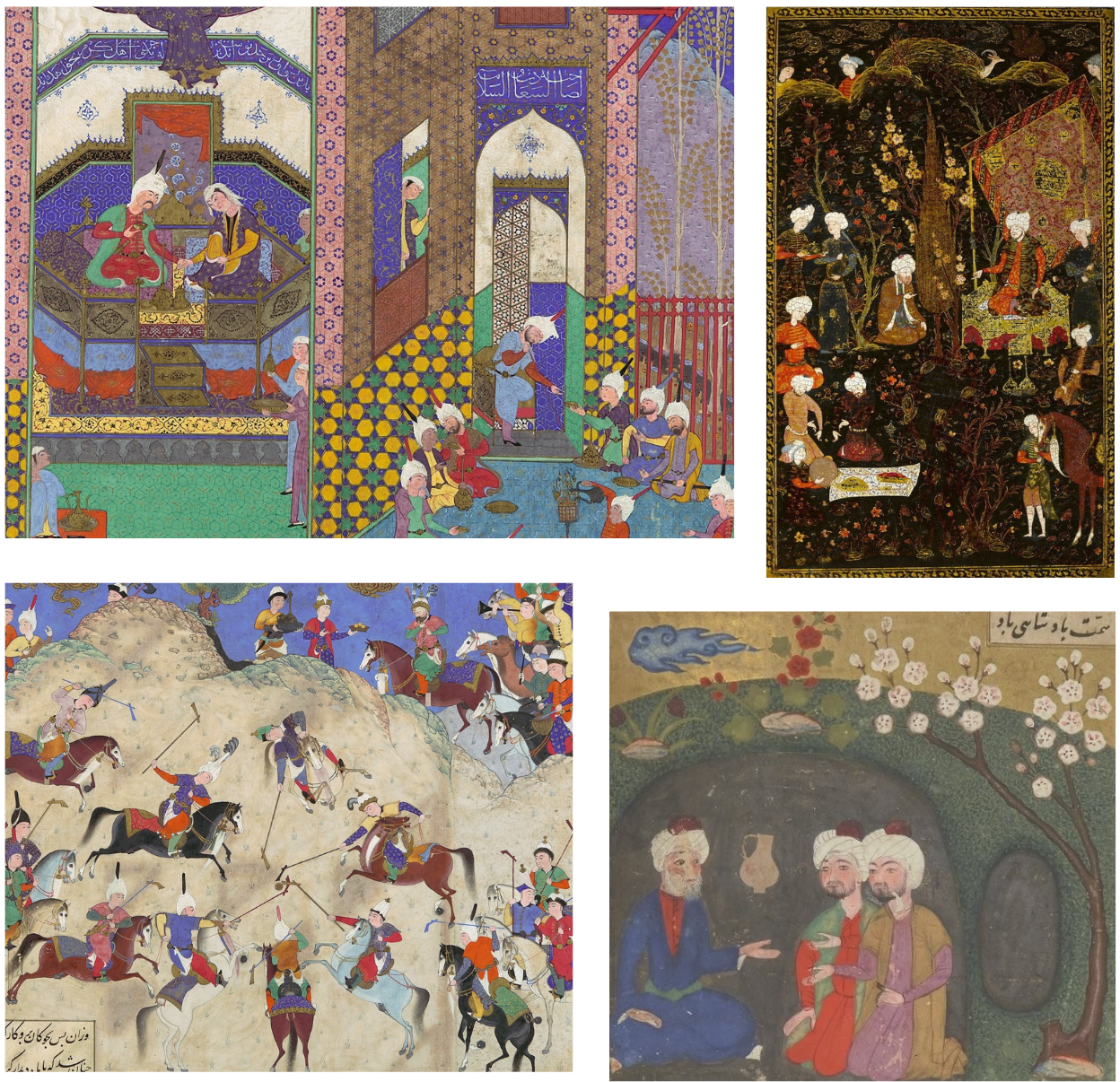}
		\label{fig:tabriz_2}
	}
	\hfill
	\subfloat[]{
		\includegraphics[width=0.4\textwidth]{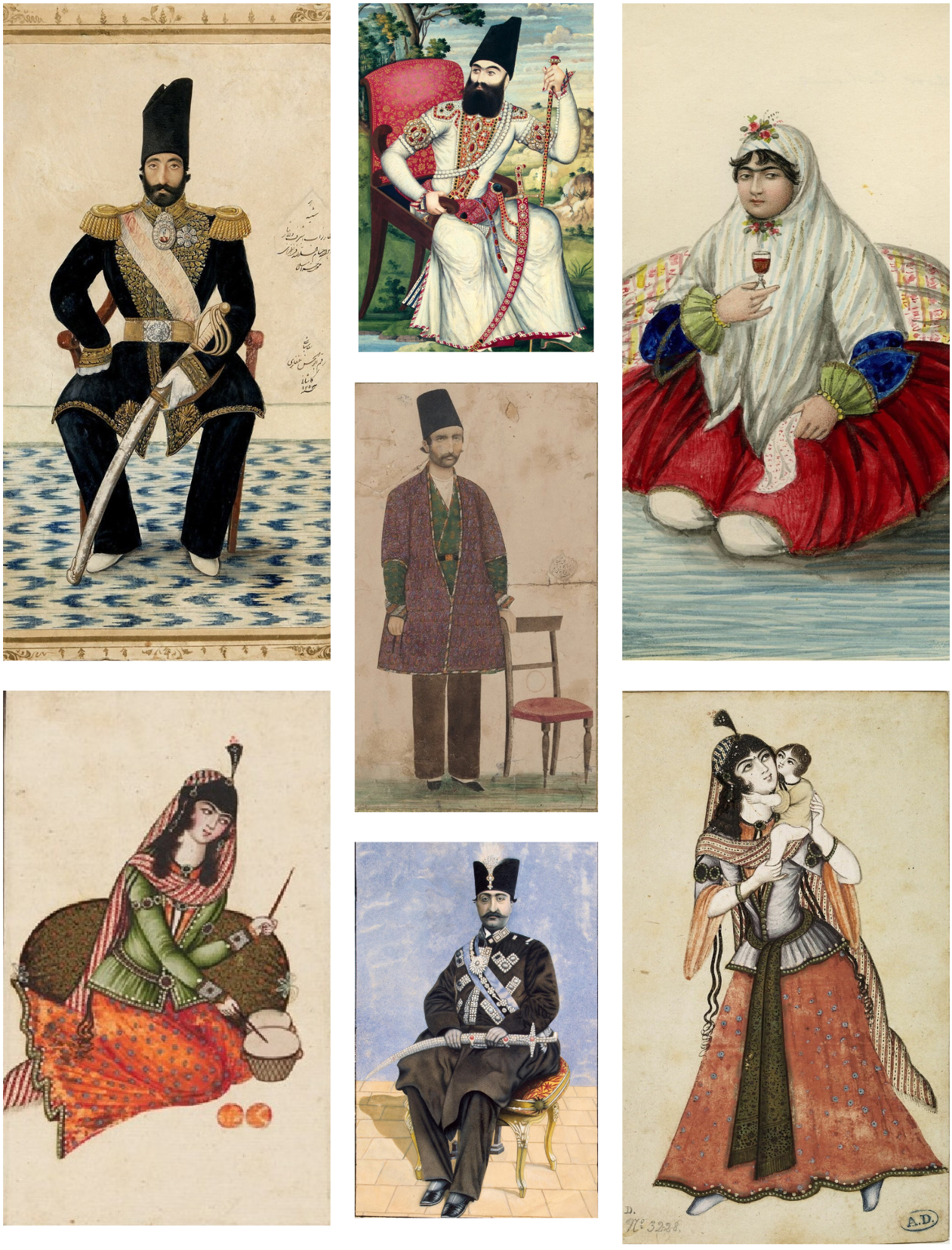}
		\label{fig:qajar}
	}
	\caption{Illustrative overview of five schools of art used in this project: (a) Herat School; (b) Shiraz-e Avval School; (c) Tabriz-e Avval School; (d) Tabriz-e Dovvom School; (e) Qajar School.}
	
	\label{fig:dataset_images}
	
\end{figure*}

\subsection{Classification}
The approach in this experiment involved dividing the input image into five patches, as mentioned in the Patch Extraction subsection. The classification results of these five sections were then used for final decision-making. This method, due to its higher hardware resource utilization and complexity, presents a challenge in design and implementation. In the approach using the image patches as input, mentioned earlier, the input data is divided into five patches, and each patch is examined separately. For each image patch, a vector with five elements (representing the class predicted by the model) is generated. Subsequently, the vectors resulting from each of these five image patches are analyzed separately. To determine the final label, we sum these vectors along their corresponding elements. In other words, for each of the five classes, we calculate the sum of the corresponding values in the predicted vectors for each image segment. This process results in a final vector with five elements, where each element represents the total scores that a particular class has received from all image segments. Then, to determine the final class, the index of the element with the highest value in this final vector is selected.

\section{EXPERIMENTAL SETUP}
\label{sec:experimentalsetup}

\subsection{Dataset}
Our research utilizes data gathered from a selected collection of images from the Islamic Painted Page database. This database, encompassing a diverse array of Islamic miniature paintings, offers users the ability to search and access images based on various aspects. For the purposes of our study, we have focused on images pertaining to five specific artistic schools to enable a more thorough examination and analysis of the characteristics unique to each school.

During the course of this research, approximately 1000 images were downloaded from the database. However, to maintain the quality and accuracy of the dataset, a portion of these images, which were of low quality or did not align with our criteria for the artistic schools, were excluded from the dataset. This crucial step was taken to prevent any decline in the dataset quality. The careful selection of images and ensuring their quality was a significant part of the data preprocessing process. This ensured that high-quality and relevant data were used for training the models. Table \ref{tab:dataset-distribution} presents the distribution of images across the five artistic schools included in our dataset, while Figure \ref{fig:dataset_images} visually illustrates representative examples from each school.

\begin{table}[t]
	\caption{Distribution of classes in the Persian Miniature dataset}
	\label{tab:dataset-distribution}
	\centering
	\begin{tabular}{|c|l|c|}
		\hline
		\textbf{Class} & \textbf{Artistic School} & \textbf{Number of Images} \\ \hline
		1 & Herat & 202 \\ \hline
		2 & Qajar & 150 \\ \hline
		3 & Shiraz-e Avval & 193 \\ \hline
		4 & Tabriz-e Avval & 190 \\ \hline
		5 & Tabriz-e Dovvom & 155 \\ \hline
	\end{tabular}
\end{table}

\begin{table}[tbp]
	\centering
	\caption{Performance metrics of evaluated classification models using 5-fold cross-validation based on mean ($\mathbb{E}$), and coefficient of variation (CV). Bold values highlight the best-performing methods.}
	\label{tab:performance_metrics}
	\begin{tabular}{ccccc}
		\toprule
		\textbf{\#} & \textbf{Base Classifier} & \textbf{$\mathbb{E}$[Train Acc.]} & \textbf{$\mathbb{E}$[Test Acc.]} & \textbf{CV[Test Acc.]} \\
		\midrule
		1 & DenseNet169 & 94.66 & \textbf{87.42} & \textbf{4.8708} \\
		2 & DenseNet201 & \textbf{95.53} & 86.52 & 5.2433 \\
		3 & DenseNet121 & 89.75 & 84.16 & 6.5325 \\
		4 & Xception & 95.06 & 82.02 & 7.0251 \\
		5 & InceptionV3 & 84.83 & 76.85 & 8.4795 \\
		6 & VGG16 & 74.33 & 74.49 & 8.8541 \\
		7 & VGG19 & 71.07 & 71.57 & 8.0447 \\
		\bottomrule
	\end{tabular}
\end{table}

In the collection of images used in this study, some of the pictures contained margins unrelated to the main content of the artwork. These margins might include areas outside the primary frame of the work used for preservation or standard photography. Therefore, to maintain the accuracy and quality of the data in analysis and classification, it was imperative to remove these irrelevant margins. In this regard, all 890 remaining images were carefully reviewed, and unnecessary margins were manually cropped out. This process helped us focus solely on the significant and relevant artistic elements of each painting school, training our machine learning model more efficiently. The Persian Miniature dataset has been made publicly available on Kaggle and can be accessed at the following link: \href{https://www.kaggle.com/datasets/mojtabashahi/persian-miniature}{https://www.kaggle.com/datasets/mojtabashahi/persian-miniature}.

\subsection{Classification Model}
In our endeavor to create a highly efficient and precise classification system for the analysis of Persian miniature paintings, our research has led us to select DenseNet169 and DenseNet201 as our primary models. This decision was informed by a comprehensive series of experiments aimed at evaluating the performance of various pre-trained deep learning models on our dataset, the results of which are summarized in Table \ref{tab:performance_metrics}. The distinguishing factor of DenseNet169 and DenseNet201 in the context of our research lies in their architectural efficiency and their demonstrated capability to handle the full complexity of the images. Unlike approaches that rely solely on analyzing segmented patches, these models excel in processing and classifying fused images.

By leveraging the strengths of DenseNet169 and DenseNet201, as evidenced by their superior performance metrics detailed in Table \ref{tab:performance_metrics}, we aim to push the boundaries of automated classification in the realm of Persian miniatures. This approach offers new insights and tools for art historians, curators, and enthusiasts alike, promising a significant advancement in the field of digital art analysis.

The DenseNet architecture, known for its densely connected convolutional networks, offers several advantages in image classification tasks. Its design promotes feature reuse and mitigates the problem of vanishing gradients, making it particularly suitable for our dataset with intricate artistic details. In our experimental trials, DenseNet169 and DenseNet201 consistently outperformed other models. This robust performance across different image formats reinforced our decision to employ these models as the backbone of our classification framework. By harnessing the strengths of DenseNet169 and DenseNet201, we aim to achieve a detailed understanding of the diverse artistic schools represented in our dataset, thus pushing the boundaries of automated Persian miniature art classification.

The weights of these pre-trained models are set to be non-trainable to ensure they remain unchanged during the training process. Instead, several new layers are added to the model to adapt it to our specific task. These layers include a GlobalAveragePooling2D layer, a Dense layer with 32 neurons and 'relu' activation, and a Dropout layer with a rate of 0.3 to prevent overfitting. The final layer, responsible for classification, contains neurons equal to the number of classes and utilizes a softmax activation function. Experiments with this model configuration were conducted on the data with a batch size of 32 for 15 epochs, utilizing Google Colab's T4 GPUs to ensure robust computational efficiency and reproducibility of the results.

\subsection{Evaluation Metrics}
In evaluating the performance of our classification models, we have adopted a dual metric approach, focusing on both accuracy and variance. This dual approach provides a comprehensive assessment of the model's effectiveness and stability across different scenarios.

Accuracy, as a primary metric, offers a straightforward measure of the model's ability to correctly classify the artistic schools in our dataset. It is quantified as the proportion of correct predictions out of the total number of predictions made. This metric is crucial for gauging the model's effectiveness in distinguishing between the intricate features of Persian miniature paintings. To further refine our evaluation, we have also incorporated the measurement of variance. Variance, in this context, refers to the consistency of the model's performance across different subsets of the data. 

It is particularly relevant given our experimental design, which employs a five-fold cross-validation technique. In this setup, the data is split into five distinct sets, with each set comprising 80\% of the data for training and the remaining 20\% for testing. This 20\% is varied in each iteration to ensure that the model is tested on all data points without overlap. The variance is then calculated based on the model's performance across these different subsets, providing insights into the model's reliability and robustness. The mean accuracy across these five iterations of cross-validation will be presented as the overall evaluation of the model. This approach allows us to not only assess the model's general accuracy but also to understand its performance variability, ensuring a more reliable evaluation.

\section{RESULTS AND DISCUSSION}
\label{sec:results}
The experimental results, as outlined in Tables \ref{tab:train_accuracy}, \ref{tab:test_accuracy} and \ref{tab:variance}, provide a comprehensive overview of the classification performance using DenseNet201 and DenseNet169 models. Notably, when analyzing the performance from a fused-image perspective by fusing the classifications from the five patches, the accuracy significantly improves compared to considering the patches independently. This improvement underscores the effectiveness of our proposed methodology in capturing the detailed characteristics of Persian miniature schools through a collective analysis of image segments.

\begin{table}[bp]
	\centering
	\caption{Training accuracy for patch and fused images classification.}
	\label{tab:train_accuracy}
	\begin{tabular}{cccc}
		\toprule
		\textbf{\#} & \textbf{Base Classifier} & \textbf{$\mathbb{E}$[Patch Train Acc.]} & \textbf{$\mathbb{E}$[Fused Image Train Acc.]} \\
		\midrule
		1 & DenseNet201 & 95.04 & 99.69 \\
		2 & DenseNet169 & 94.42  & 99.8  \\
		\bottomrule
	\end{tabular}
\end{table}

The training accuracy data, as presented in Table \ref{tab:train_accuracy}, reveals high performance levels for both DenseNet201 and DenseNet169 models, with the $\mathbb{E}$[Fused Image Train Acc.] reaching 99.69\% and 99.8\%, respectively. While these demonstrate the models' exceptional ability to learn and adapt to the training dataset, a comparison with the test accuracy from Table \ref{tab:test_accuracy} indicates a potential concern regarding overfitting.

As indicated in Table \ref{tab:test_accuracy}, the baseline test accuracy for individual patches using DenseNet201 and DenseNet169 models hovered around 85\%. However, when employing a patching strategy that considers the image as a fusion of its five classified patches, we observed a notable increase in accuracy. Specifically, DenseNet201 exhibited an enhanced test accuracy of 91.69\%, and DenseNet169 showed a similar improvement, reaching 89.78\%. This uplift in accuracy highlights the synergistic effect of analyzing multiple image segments together, allowing for a more holistic and accurate classification of the Persian miniatures.

The Coefficient of Variation (CV) data presented in Table \ref{tab:variance} offers insights into the stability of the models across different test scenarios. For DenseNet201, the CV remained approximately constant even after the fusion of patch classifications. This consistency suggests that DenseNet201 is robust to variations in the dataset and maintains its performance reliability across different image compositions. On the other hand, DenseNet169, while showing improved accuracy through fusion, exhibited a slight increase in CV. This suggests that DenseNet169 may be slightly more sensitive to dataset variations compared to DenseNet201, potentially leading to less consistent performance across different test sets.

The findings from our study have significant implications for the field of digital art analysis, particularly in the classification of Persian miniatures. The improved accuracy achieved through the fusion of classified patches underscores the potential of machine learning techniques in understanding and categorizing complex artistic styles. Moreover, the comparative stability of DenseNet201 suggests it as a preferable model for researchers and practitioners aiming for consistent classification performance in similar tasks.

\begin{table}[t]
	\centering
	\caption{Comparison of average test accuracy for patch and fused images classification.}
	\label{tab:test_accuracy}
	\begin{tabular}{cccccc}
		\toprule
		\textbf{\#} & \textbf{Base Classifier}  & \textbf{$\mathbb{E}$[Patch Test Acc.]}  & \textbf{$\mathbb{E}$[Fused Image Test Acc.]} \\
		\midrule
		1 & DenseNet201  & 86.67  & 91.69 \\
		2 & DenseNet169  & 84.58  & 89.78 \\
		\bottomrule
	\end{tabular}
\end{table}

\begin{table}[t]
	\centering
	\caption{Coefficient of variation analysis (expressed as percentages) for patch and fused images, based on test accuracy results.}
	\label{tab:variance}
	\begin{tabular}{cccc}
		\toprule
		\textbf{\#} & \textbf{Base Classifier} & \textbf{CV[Patch Test Acc.]} & \textbf{CV[Fused Image Test Acc.]} \\
		\midrule
		1 & DenseNet201 & 4.12 & 4.09 \\
		2 & DenseNet169 & 2.95 & 3.84 \\
		\bottomrule
	\end{tabular}
\end{table}

\begin{figure}[t]
	\centering
	\includegraphics[width=0.5\textwidth]{"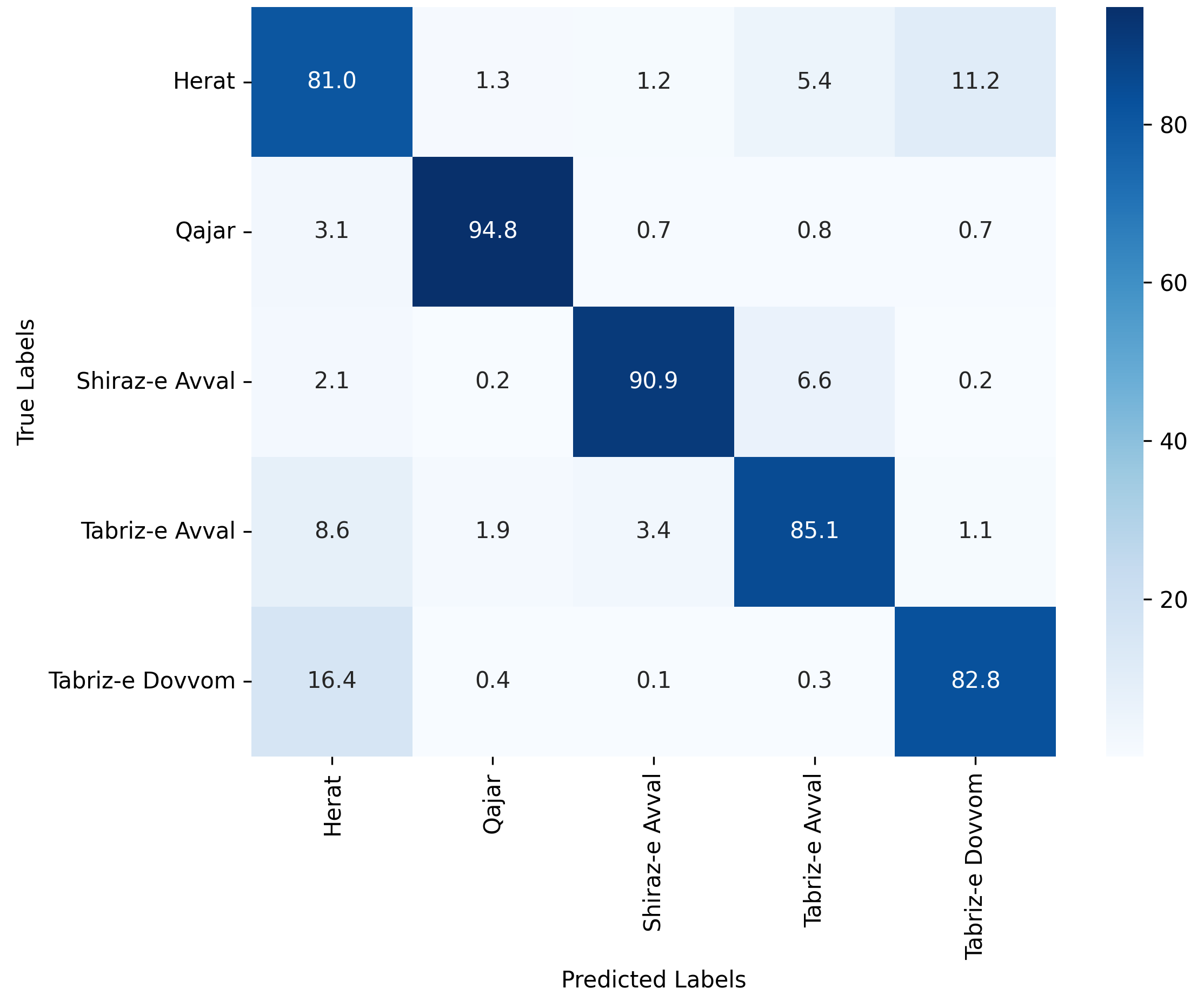"}
	\caption{Normalized Confusion Matrix of Patch-Level Classification using DenseNet201: Mean Across 5-Fold Cross-Validation.}
	\label{fig:patches_cm_mean_normalized_5foldCV_DenseNet201}
\end{figure}

The confusion matrix for the patch analysis using the DenseNet201 model, as visualized in Figure \ref{fig:patches_cm_mean_normalized_5foldCV_DenseNet201}, encapsulates the model's performance on a segmental level through a 5-fold cross-validation. This granular analysis, based on the classification of individual image patches, highlights the model's capacity to discern and categorize the unique features present in different segments of Persian miniatures. Notwithstanding a high degree of accuracy in certain classes, the matrix also reveals instances of misclassification, suggesting areas where model optimization may yield improvements. Such discrepancies may reflect the inherent limitations of partial data representation in the patches, which could lack the contextual richness necessary for accurate full image classification.

\begin{figure}[t]
	\centering
	\includegraphics[width=0.5\textwidth]{"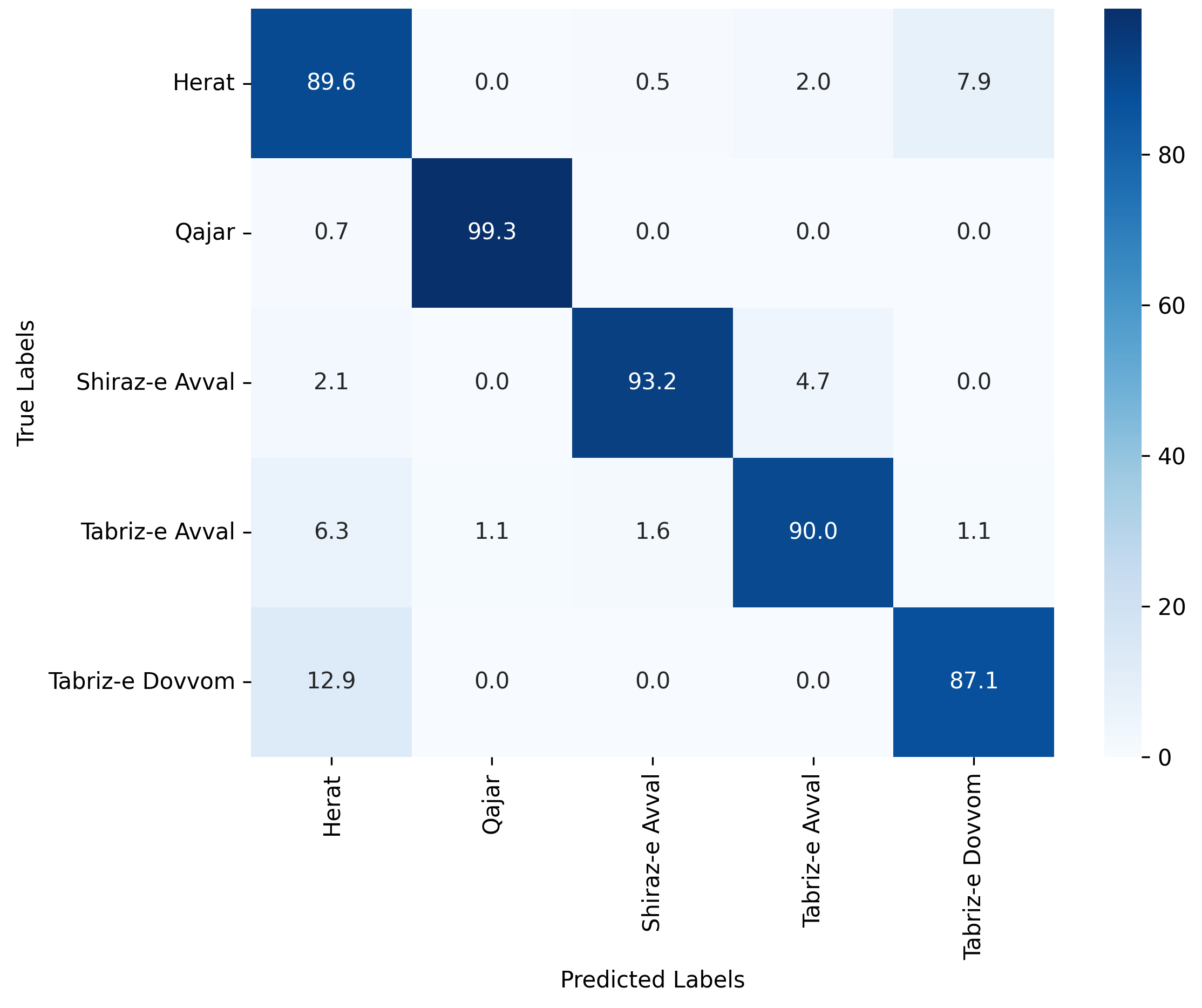"}
	\caption{Normalized Confusion Matrix of Fused-Image Classification using DenseNet201: Mean Across 5-Fold Cross-Validation.}
	\label{fig:cm_mean_normalized_5foldCV_DenseNet201}
\end{figure}

Conversely, the confusion matrix for the fused image analysis, depicted in Figure \ref{fig:cm_mean_normalized_5foldCV_DenseNet201}, conveys a substantially different outcome. By integrating the classifications from the five patches to represent the complete image, the DenseNet201 model exhibits a notable enhancement in overall accuracy. This synthesized perspective aligns more closely with traditional methods of art analysis, wherein the artwork is evaluated in its entirety. The resulting lower misclassification rates and reduced variance underscore the advantages of comprehensive image analysis. This model's ability to effectively combine discrete patch data into a unified classification reaffirms the importance of considering the full canvas of Persian miniatures to capture the depth and nuance of their artistic heritage.

\section{CONCLUSION}
\label{sec:conclusion}

Our research embarked on addressing the intricate challenge of classifying Persian miniatures into their distinct schools of art, a task that not only demands precision but also a deep understanding of distinguishing one school from another. Despite the rich cultural heritage encapsulated in these miniatures, their classification has remained a largely manual and subjective process, underscoring the need for a more systematic and accurate method to appreciate and preserve their legacy. Our research demonstrates the effectiveness of employing a patching strategy for classifying Persian miniatures into their respective schools of art, achieving an average accuracy rate of over 91\%. By analyzing the image as a fusion of classified patches, we achieved higher accuracy rates, thereby enhancing the model's ability to capture the intricate details and stylistic subtleties of Persian miniatures. The comparative analysis of DenseNet201 and DenseNet169 models further provides valuable insights into model selection based on performance stability and accuracy. This study sets the stage for extensive future work in the realm of machine learning and art history, specifically regarding Persian miniatures. Future endeavors could include expanding the scope of the research to encompass a broader range of art schools, thereby enriching the dataset and potentially unveiling new insights into stylistic evolution and influences across different periods and regions. Furthermore, integrating more advanced machine learning techniques could refine the accuracy of classification and offer deeper understandings of the intricate artistry of Persian miniatures. Through such expansions, this line of research can significantly contribute to the digital preservation and study of cultural heritage, opening new avenues for exploration in the intersection of technology, art, and history.






\bibliographystyle{agsm}
\bibliography{references}

@article{qizilbash_hat,
	title={Invisible Violence In Persian Painting},
	author={Moghaddam, Visheh Khatami},
	journal={International Journal of {\v{Z}}i{\v{z}}ek Studies},
	volume={15},
	number={3},
	year={2021}
}

@article{yosinski2014transferable,
	title={How transferable are features in deep neural networks?},
	author={Yosinski, Jason and Clune, Jeff and Bengio, Yoshua and Lipson, Hod},
	journal={Advances in neural information processing systems},
	volume={27},
	year={2014}
}

@article{srivastava2014dropout,
	title={Dropout: a simple way to prevent neural networks from overfitting},
	author={Srivastava, Nitish and Hinton, Geoffrey and Krizhevsky, Alex and Sutskever, Ilya and Salakhutdinov, Ruslan},
	journal={The journal of machine learning research},
	volume={15},
	number={1},
	pages={1929--1958},
	year={2014},
	publisher={JMLR. org}
}

@book{o2021studies,
	title={Studies in Islamic Painting, Epigraphy and Decorative Arts},
	author={O’Kane, Bernard},
	year={2021},
	publisher={Edinburgh University Press}
}

@book{loukonine2010persian,
	title={Persian Miniatures},
	author={Loukonine, V. and Lukonin, V.G. and Ivanov, A.A.},
	isbn={9781844847822},
	series={Mega Square Series},
	url={https://books.google.com/books?id=LxuBQgAACAAJ},
	year={2010},
	publisher={Parkstone International}
}

@inproceedings{zujovic2009classifying,
	title={Classifying paintings by artistic genre: An analysis of features \& classifiers},
	author={Zujovic, Jana and Gandy, Lisa and Friedman, Scott and Pardo, Bryan and Pappas, Thrasyvoulos N},
	booktitle={2009 IEEE International Workshop on Multimedia Signal Processing},
	pages={1--5},
	year={2009},
	organization={IEEE}
}

@inproceedings{xiao2010sun,
	title={Sun database: Large-scale scene recognition from abbey to zoo},
	author={Xiao, Jianxiong and Hays, James and Ehinger, Krista A and Oliva, Aude and Torralba, Antonio},
	booktitle={2010 IEEE computer society conference on computer vision and pattern recognition},
	pages={3485--3492},
	year={2010},
	organization={IEEE}
}

@article{oliva2001modeling,
	title={Modeling the shape of the scene: A holistic representation of the spatial envelope},
	author={Oliva, Aude and Torralba, Antonio},
	journal={International journal of computer vision},
	volume={42},
	pages={145--175},
	year={2001},
	publisher={Springer}
}

@inproceedings{dalal2005histograms,
	title={Histograms of oriented gradients for human detection},
	author={Dalal, Navneet and Triggs, Bill},
	booktitle={2005 IEEE computer society conference on computer vision and pattern recognition (CVPR'05)},
	volume={1},
	pages={886--893},
	year={2005},
	organization={Ieee}
}

@inproceedings{lazebnik2006beyond,
	title={Beyond bags of features: Spatial pyramid matching for recognizing natural scene categories},
	author={Lazebnik, Svetlana and Schmid, Cordelia and Ponce, Jean},
	booktitle={2006 IEEE computer society conference on computer vision and pattern recognition (CVPR'06)},
	volume={2},
	pages={2169--2178},
	year={2006},
	organization={IEEE}
}

@inproceedings{ahonen2009rotation,
	title={Rotation invariant image description with local binary pattern histogram fourier features},
	author={Ahonen, Timo and Matas, Ji{\v{r}}{\'\i} and He, Chu and Pietik{\"a}inen, Matti},
	booktitle={Image Analysis: 16th Scandinavian Conference, SCIA 2009, Oslo, Norway, June 15-18, 2009. Proceedings 16},
	pages={61--70},
	year={2009},
	organization={Springer}
}

@article{matas2004robust,
	title={Robust wide-baseline stereo from maximally stable extremal regions},
	author={Matas, Jiri and Chum, Ondrej and Urban, Martin and Pajdla, Tom{\'a}s},
	journal={Image and vision computing},
	volume={22},
	number={10},
	pages={761--767},
	year={2004},
	publisher={Elsevier}
}

@inproceedings{sivic2004video,
	title={Video data mining using configurations of viewpoint invariant regions},
	author={Sivic, Josef and Zisserman, Andrew},
	booktitle={Proceedings of the 2004 IEEE Computer Society Conference on Computer Vision and Pattern Recognition, 2004. CVPR 2004.},
	volume={1},
	pages={I--I},
	year={2004},
	organization={IEEE}
}

@inproceedings{shechtman2007matching,
	title={Matching local self-similarities across images and videos},
	author={Shechtman, Eli and Irani, Michal},
	booktitle={2007 IEEE Conference on Computer Vision and Pattern Recognition},
	pages={1--8},
	year={2007},
	organization={IEEE}
}

@article{torralba200880,
	title={80 million tiny images: A large data set for nonparametric object and scene recognition},
	author={Torralba, Antonio and Fergus, Rob and Freeman, William T},
	journal={IEEE transactions on pattern analysis and machine intelligence},
	volume={30},
	number={11},
	pages={1958--1970},
	year={2008},
	publisher={IEEE}
}

@inproceedings{kovsecka2002video,
	title={Video compass},
	author={Ko{\v{s}}eck{\'a}, Jana and Zhang, Wei},
	booktitle={Computer Vision—ECCV 2002: 7th European Conference on Computer Vision Copenhagen, Denmark, May 28--31, 2002 Proceedings, Part IV 7},
	pages={476--490},
	year={2002},
	organization={Springer}
}

@inproceedings{martin2001database,
	title={A database of human segmented natural images and its application to evaluating segmentation algorithms and measuring ecological statistics},
	author={Martin, David and Fowlkes, Charless and Tal, Doron and Malik, Jitendra},
	booktitle={Proceedings Eighth IEEE International Conference on Computer Vision. ICCV 2001},
	volume={2},
	pages={416--423},
	year={2001},
	organization={IEEE}
}

@article{hoiem2007recovering,
	title={Recovering surface layout from an image},
	author={Hoiem, Derek and Efros, Alexei A and Hebert, Martial},
	journal={International Journal of Computer Vision},
	volume={75},
	pages={151--172},
	year={2007},
	publisher={Springer}
}

@article{lalonde2007photo,
	title={Photo clip art},
	author={Lalonde, Jean-Fran{\c{c}}ois and Hoiem, Derek and Efros, Alexei A and Rother, Carsten and Winn, John and Criminisi, Antonio},
	journal={ACM transactions on graphics (TOG)},
	volume={26},
	number={3},
	pages={3--es},
	year={2007},
	publisher={ACM New York, NY, USA}
}

@article{blessing2010using,
	title={Using machine learning for identification of art paintings},
	author={Blessing, Alexander and Wen, Kai},
	journal={Technical report},
	year={2010},
	publisher={Stanford University}
}

@inproceedings{folego2016impressionism,
	title={From impressionism to expressionism: Automatically identifying van Gogh's paintings},
	author={Folego, Guilherme and Gomes, Otavio and Rocha, Anderson},
	booktitle={2016 IEEE international conference on image processing (ICIP)},
	pages={141--145},
	year={2016},
	organization={IEEE}
}

@inproceedings{levy2014genetic,
	title={Genetic algorithms and deep learning for automatic painter classification},
	author={Levy, Erez and David, Omid E and Netanyahu, Nathan S},
	booktitle={proceedings of the 2014 Annual Conference on Genetic and Evolutionary Computation},
	pages={1143--1150},
	year={2014}
}

@inproceedings{david2016deeppainter,
	title={Deeppainter: Painter classification using deep convolutional autoencoders},
	author={David, Omid E and Netanyahu, Nathan S},
	booktitle={Artificial Neural Networks and Machine Learning--ICANN 2016: 25th International Conference on Artificial Neural Networks, Barcelona, Spain, September 6-9, 2016, Proceedings, Part II 25},
	pages={20--28},
	year={2016},
	organization={Springer}
}

@article{sandoval2019two,
	title={Two-stage deep learning approach to the classification of fine-art paintings},
	author={Sandoval, Catherine and Pirogova, Elena and Lech, Margaret},
	journal={IEEE Access},
	volume={7},
	pages={41770--41781},
	year={2019},
	publisher={IEEE}
}

%
%
%
\end{document}